# HiLa: Hierarchical Vision-Language Collaboration for Cancer Survival Prediction

Jiaqi Cui[1], Lu Wen[1], Yuchen Fei[1], Bo Liu[2], Luping Zhou[3], Dinggang Shen[4,5,6], Yan Wang[1(✉)]

[1] School of Computer Science, Sichuan University, Chengdu, China
wangyanscu@hotmail.com
[2] Department of Computing, The Hong Kong Polytechnic University, Hong Kong, China
[3] School of Electrical and Computer Engineering, University of Sydney, Sydney, Australia
[4] School of Biomedical Engineering & State Key Laboratory of Advanced Medical Materials and Devices, ShanghaiTech University, Shanghai, China
[5] Shanghai United Imaging Intelligence Co., Ltd., Shanghai, China
[6] Shanghai Clinical Research and Trial Center, Shanghai, China

**Abstract.** Survival prediction using whole-slide images (WSIs) is crucial in cancer research. Despite notable success, existing approaches are limited by their reliance on sparse slide-level labels, which hinders the learning of discriminative representations from gigapixel WSIs. Recently, vision language (VL) models, which incorporate additional language supervision, have emerged as a promising solution. However, VL-based survival prediction remains largely unexplored due to two key challenges. First, current methods often rely on only one simple language prompt and basic cosine similarity, which fails to learn fine-grained associations between multi-faceted linguistic information and visual features within WSI, resulting in inadequate vision-language alignment. Second, these methods primarily exploit patch-level information, overlooking the intrinsic hierarchy of WSIs and their interactions, causing ineffective modeling of hierarchical interactions. To tackle these problems, we propose a novel **Hi**erarchical vision-**La**nguage collaboration (HiLa) framework for improved survival prediction. Specifically, HiLa employs pretrained feature extractors to generate hierarchical visual features from WSIs at both patch and region levels. At each level, a series of language prompts describing various survival-related attributes are constructed and aligned with visual features via Optimal Prompt Learning (OPL). This approach enables the comprehensive learning of discriminative visual features corresponding to different survival-related attributes from prompts, thereby improving vision-language alignment. Furthermore, we introduce two modules, i.e., Cross-Level Propagation (CLP) and Mutual Contrastive Learning (MCL) to maximize hierarchical cooperation by promoting interactions and consistency between patch and region levels. Experiments on three The Cancer Genome Atlas (TCGA) datasets demonstrate our state-of-the-art performance.

# 1    Introduction

Survival prediction is a fundamental topic in cancer research which aims to assess the relative risks of death in cancer patients [1]. Accurate survival prediction holds significant importance in supporting clinical decision-making and maximizing therapeutic benefits [2]. Digital whole slide images (WSIs), which capture microscopic changes in tumor cells and their microenvironment, have become indispensable in computational pathology (CPath) for survival prediction [3].

Given the ultra-high resolution and expensive pixel-level annotations of WSIs, multiple instance learning (MIL) paradigm has become the predominant approach for WSI analysis. Traditional MIL-based methods [4-12], which rely solely on slide-level annotation for supervision, typically follow a three-step process: 1) dividing a WSI (slide) into multiple small patches (instances); 2) extracting visual features from these instances using a pretrained encoder; and 3) aggregating the instances features for slide-level prediction. Despite achieving "clinical-grade" performance, these methods are constrained by their exclusive dependence on sparse slide-level labels, making it challenging to learn sufficient survival-related discriminative visual features from gigapixel WSIs. Recently, vision-language (VL) models have shown impressive performance across various tasks [13-18] by incorporating additional language supervision, providing a promising solution to the above-mentioned limitation of the MIL paradigm in CPath. For example, Zhang *et al.* [19] utilized disease-level biomedical description to enhance the interpretation of WSIs. Qu *et al.* [20] introduced pathology language prompts at for improved WSI classification. More recently, Liu *et al.* [21] presented the first vision-language framework for survival prediction, which leverages prognostic language priors to assign weights to different instances during aggregation.

Despite notable progress, designing effective VL-based methods for survival prediction in CPath remains an open topic. The first challenge lies in the *insufficient vision-language alignment*. In clinic, pathologists rely on multiple cues (also regarded as language prompts) to assess patient survival from various perspectives (tumor differentiation, invasion, and lymph node metastasis, *etc.*). However, existing methods either condense this rich information to one oversimplified prompt (e.g., an H&E image of {cancer type}) [19, 22] or match all prompts with visual features using basic cosine similarity [21, 23]. This rudimentary alignment paradigm fails to capture fine-grained associations between multi-faceted pathological linguistic information and complex visual features within WSIs. The second challenge stems from the *ineffective modeling of hierarchical interactions*. WSIs, at a given magnification, exhibit a hierarchy across varying image resolutions, i.e., lower resolutions (e.g., 256×256 patch level) capture local cellular morphology, whereas higher resolutions (e.g., 4096×4096 region level) reveal global tissue organizations and tumor-immune interactions. However, current VL-based survival prediction methods primarily focus on patch-level analysis. This narrow focus has risk of overlooking critical global contexts, such as broader spatial organization of phenotypes in microenvironment, limiting prediction accuracy.

Motivated to address the above limitations, in this paper, we propose a novel **Hi**erarchical vision-**La**nguage collaboration (HiLa) framework for cancer survival prediction. Specifically, our network employs pretrained feature extractors [24] to produce hierarchical visual features from WSIs at both local patch and global region levels. At each level, we first utilize a large language model (LLM) to generate multiple language prompts that highlight diverse survival-related attributes. Then, inspired by [25, 26], we adopt Optimal Prompt Learning (OPL) to establish the best correspondence between diverse attributes within different prompts and WSI visual features, thereby facilitating the comprehensive extraction of discriminative visual features associated with survival assessment. On the other hand, to enhance interactions between patch level and region level, we design a Cross-Level Propagation (CLP) module that leverages patch-level knowledge to support region-level prediction, establishing a hierarchical and attentive connection between the two levels. Furthermore, we introduce a Mutual Contrastive Learning (MCL) module to enforce consistency between patch-level and region-level visual features for each patient, further boosting hierarchical interactions. Through the above hierarchical vision-language collaboration, our HiLa fully exploits the model's representation capability, leading to improved survival prediction accuracy.

The contributions of this paper are summarized as follows. (1) We propose a novel hierarchical vision-language collaboration framework for survival prediction in CPath. (2) We introduce multiple language prompts and employ OPL to enhance the alignment between diverse survival-related attributes in the prompt set and WSI visual features, thereby improving the extraction of discriminative visual patterns. (3) We develop two modules, namely CLP and MCL, to ensure effective hierarchical interaction and consistency. (4) Experimental results on three public cancer datasets from The Cancer Genome Atlas (TCGA) verify our superiority.

## 2　Methodology

### 2.1　Overview and Problem Formulation

The overall framework of the proposed method is illustrated in Fig. 1. Our network extracts visual features (tokens) from a WSI at both patch and region levels, capturing local cellular morphology and global tissue organization, respectively. Meanwhile, survival-related sentences are generated and encoded as a set of language prompts at both levels to guide the selection and aggregation of discriminative visual tokens via hierarchical vision-language collaboration. Details are provided in the following subsections.

**Hierarchical Bags Construction.** To capture the intrinsic hierarchy of WSIs, we employ pretrained feature extractors [24], which consist of a patch-level extractor $E_P(\cdot)$ and a region-level extractor $E_R(\cdot)$. Given the set of WSIs at 20× magnification, $E_P(\cdot)$ and $E_R(\cdot)$ generate hierarchical bags $H_P = \{H_P^i\}_{i=1}^{M_P} \in \mathbb{R}^{M_P \times d}$ for patches of 256×256 and $H_R = \{H_R^i\}_{i=1}^{M_R} \in \mathbb{R}^{M_R \times d}$ for regions of 4096×4096, respectively, where $H_P^i$ and $H_R^i$ denote visual tokens. $M_P$ and $M_R$ represent the total numbers of extracted patch-level and region-level visual tokens, respectively; $d$ is the corresponding channel dimension.

**Language Prompts Generation.** We generate a set of language prompts that describe various perspectives on survival outcomes for comprehensive learning. To this end, we first construct a query for each cancer type: *"What visual features of H&E-stained pathological images are related to the survival prediction of {cancer type} at the*

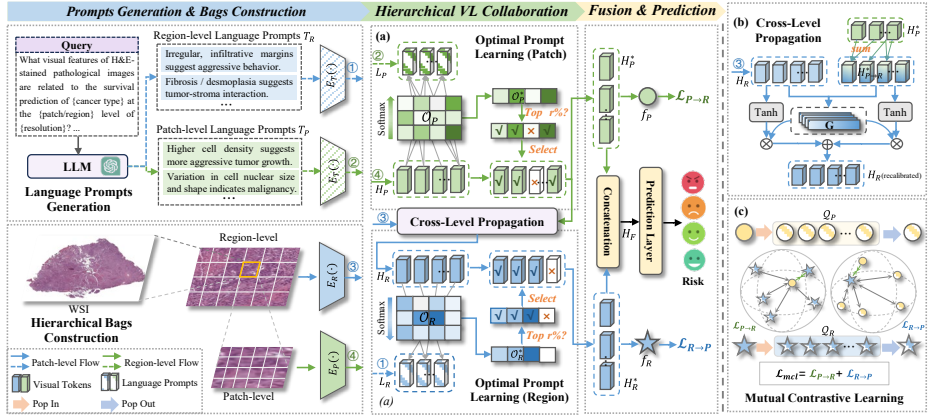

**Fig. 1.** Overview of the proposed HiLa. (a) The OPL module aligns language prompts with visual to-kens extracted from WSIs. (b) The CLP and (c) the MCL modules facilitate effective interactions between patch-level and region-level learning, enhancing hierarchical cooperation. Finally, visual tokens learned from both levels are concatenated to estimate risk scores.

*{patch/region} level of {resolution}? Please list them point by point."* The query is input into LLM (i.e., GPT-4o) to generate multiple sentences $T_P = \{T_P^i\}_{i=1}^{N_P}$ and $T_R = \{T_R^i\}_{i=1}^{N_R}$ that detail different survival-related attributes observable in WSIs at the patch and region levels, respectively, where $N_P$ and $N_R$ are the number of sentences. The generated sentences $T_P$ and $T_R$ are then processed by PLIP's [3] text encoder $E_T(\cdot)$ to produce the encoded language prompts, denoted as $L_P = \{L_P^i\}_{i=1}^{N_P \times d}$ and $L_R = \{L_R^i\}_{i=1}^{N_R \times d}$.

**Survival Prediction Formulation.** Survival prediction estimates the risk probability of an event before a specific time. Let $\mathcal{D} = \{P^n, c^n, t^n\}_{n=1}^N$ represent the data of $N$ patients, where $P^n$ is WSIs for the $n$-th patient, $c^n \in \{0,1\}$ is the right censor status, and $t^n$ is the overall survival time. Given $P^n$, we can acquire patch-level token bag $H_P^n$ and region-level token bag $H_R^n$, and develop their associated sets of language prompts $L_P^n$ and $L_R^n$. Then, discriminative visual tokens are extracted via hierarchical vision-language collaboration, yielding the final token set, denoted as $H_F^n$. The aim of our network is to use $H_F^n$ to estimate the hazard function $\mathcal{F}_{hazard}^n(t|H_F^n)$, which is the probability of death occurring after a time point $t$. Instead of directly predicting the overall survival time $t^n$, survival function $\mathcal{F}_{surv}^n(t|H_F^n)$ estimates the risk by cumulating the negated hazard functions, defined as:

$$\mathcal{F}_{surv}^n(t|H_F^n) = \prod_{s=1}^t (1 - \mathcal{F}_{hazard}^n(s|H_F^n)), \tag{1}$$

### 2.2    Hierarchical Vision-Language Collaboration

In hierarchical vision-language collaboration, we adopt optimal prompt learning (OPL) to align diverse survival-related attributes in language prompts $L_{[P,G]}$ with visual tokens $H_{[P,G]}$ at both the patch and region levels. Moreover, a Cross-Level Propagation (CLP) module bridges the two levels, enabling patch-level knowledge to support region-level predictions. During this process, a Mutual Contrastive Learning (MCL) module is incorporated to ensure the consistency between predictions at the two levels.

**Optimal Prompt Learning.** The OPL is designed to enhance the alignment between multiple language prompts and visual tokens, thereby enhancing the comprehensive learning of discriminative visual tokens corresponding to different survival-related attributes within gigapixel WSIs. As shown in Fig. 1(a), given the visual tokens $H_{[P,R]}$ and language prompts $L_{[P,R]}$ at patch and region levels (where the superscript $n$ will be omitted when unnecessary for notational brevity), we first compute the optimal matching plan $\mathcal{O}_{[P,R]}$ with the optimal matching cost $d_{OT}$ to align $L_{[P,R]}$ to $H_{[P,R]}$, defined as:

$$d_{OT}\big(u,v|\mathcal{C}_{[P,R]}\big) = \min_{\mathcal{O}_{[P,R]}} < \mathcal{O}_{[P,R]}, \mathcal{C}_{[P,R]} >,$$
$$s.t.\ \mathcal{O}_{[P,R]}1_{M_{[P,R]}} = u,\quad \mathcal{O}_{[P,R]}1_{N_{[P,R]}} = v,\quad \mathcal{O}_{[P,R]} \in \mathbb{R}_+^{M_{[P,R]} \times N_{[P,R]}}, \tag{2}$$

where $\mathcal{C}_{[P,R]} = 1 - sim\,(H_{[P,R]}, L_{[P,R]})$ is the cost matrix that quantifies the similarity between each token in $H_{[P,R]}$ and each prompt in $L_{[P,R]}$; $u$ and $v$ are marginal distributions. The obtained $\mathcal{O}_{[P,R]}$ assesses correspondence degree between language prompts and visual tokens, ensuring that each language prompt is optimally matched to the most relevant visual tokens. Then, $\mathcal{O}_{[P,R]}$ is converted to optimal matching probability $\mathcal{O}'_{[P,R]}$:

$$\mathcal{O}'_{[P,R]}[m,n] = \frac{\exp\big(1 - \mathcal{O}_{[P,R]}[m,n]\big)}{\sum_{m'=1}^{M}\exp\big(1 - \mathcal{O}_{[P,R]}[m',n]\big)} \in \mathbb{R}^{M_{[P,R]} \times N_{[P,R]}}, \tag{3}$$

where $m \in \{1,\dots,M_{[P,R]}\}$ and $n \in \{1,\dots,N_{[P,R]}\}$. Next, we sum $\mathcal{O}'_{[P,R]}$ along the second (language) dimension, thus obtaining the optimal alignment score $\mathcal{O}^*_{[P,R]} \in \mathbb{R}^{M_{[P,R]}}$. Each element in $\mathcal{O}^*_{[P,R]}$ measures the alignment between a visual token and all survival-related attributes within language prompts. Visual tokens with higher matching probabilities are considered more related to survival assessment. Accordingly, we reorganize the visual tokens $H_{[P,R]}$ by ranking $\mathcal{O}^*_{[P,R]}$ in descending order, and selecting the top $r\%$ visual tokens, yielding $H^*_{[P,R]} \in \mathbb{R}^{(M_{[P,R]} \times r\%) \times d}$ that contains rich discriminative information related to patient survival.

**Cross-Level Propagation (CLP) Module.** We develop the CLP module which establishes an attentive and hierarchical connection between the patch and region levels, enabling patch-level knowledge to effectively guide region-level learning, as displayed in Fig. 1(b). Instead of naive concatenation or summation, CLP uses a gating mechanism to adaptively modulate the influence of patch-level tokens to region-level predictions. As each region corresponds to multiple patch-level visual tokens, the above selected $H^*_P$ are summed according to their belonged regions, yielding the selected region-level inputs, denoted as $H^*_{P \to R}$. We concatenate $H^*_{P \to R}$ with the unselected region-level inputs $H_R$ to generate a gate matrix $G$ via linear projection, followed by a sigmoid activation function. Then, $G$ is employed to adjust the weights of Tanh-activated $H^*_{P \to R}$ and $H_R$. The above process can be formulated as follows:

$$G = \text{Sigmoid}(\text{Concat}(H^*_{P \to R}; H_R); \theta_G),$$
$$H_R = G * \text{Tanh}(H^*_{P \to R}; \theta_P) + (1 - G) * \text{Tanh}(H_R; \theta_R), \tag{4}$$

where $\{\theta_G, \theta_P, \theta_R\}$ are learnable parameters. The obtained recalibrated $H_R$ is then used as the input to the OPL module at the region level, enabling more precise predictions.

**Mutual Contrastive Learning (MCL) Module.** As illustrated in Fig. 1(c), we introduce a MCL module to maintain consistency between the selected patch-level and region-level visual tokens for each patient, further improving hierarchical cooperation. Concretely, given $H^*_{[P,R]}$, we compute prototypes $f_{[P,R]} \in \mathbb{R}^{(M_{[P,R]} \times r\%) \times 1}$ by summing across their channel dimension. During the training, we initialize a patch-level memory queue $\mathcal{Q}_P$ with $B-1$ patch prototypes, and a region-level memory queue $\mathcal{Q}_R$ with the same number of region prototypes. In subsequent iterations, $\mathcal{Q}_{[P,R]}$ are updated by storing the current $f_{[P,R]}$ and popping out the oldest entries. The MCL module performs contrastive learning at both the patch and region levels. At the patch level, denote the current patch prototype $f_P^n$ (for the $n$-th patient) as the anchor sample, while the corresponding region prototype $f_R^n$ (for the same patient) acts as the positive sample; and the other region prototypes (from different patients) in $\mathcal{Q}_R$ are regarded as negative samples. Based on this, the patch-level contrastive loss $\mathcal{L}_{P \rightarrow R}$ for $N$ patients is defined as:

$$\mathcal{L}_{P \rightarrow R} = -\sum_{n=1}^{N} log \frac{exp(f_P^n \cdot f_R^n)}{exp(f_P^n \cdot f_R^n) + \sum_{k=1, k \neq n}^{B-1} exp(f_P^n \cdot f_R^k)}, \quad (5)$$

Similarly, we can derive the region-level contrastive loss $\mathcal{L}_{R \rightarrow P}$. Together, these two losses constitute the MCL loss, which is expressed as:

$$\mathcal{L}_{mcl} = \mathcal{L}_{P \rightarrow R} + \mathcal{L}_{R \rightarrow P}, \quad (6)$$

### 2.3 Fusion and Prediction

To yield the final survival prediction, we concatenate the selected patch-level tokens $H_P^*$ and region-level tokens $H_R^*$ to form the final token set $H_F^n$. Then, $H_F^n$ is passed through a linear-based prediction layer to estimate risk scores. The negative log-likelihood (NLL) loss [27] is applied to supervise this process, which is defined as follows:

$$\mathcal{L}_{sur} = \sum_{n=1}^{N} -c^n log\big(\mathcal{F}_{surv}^n(t|H_F^n)\big) - (1-c^n) \log\big(\mathcal{F}_{surv}^n(t-1|H_F^n)\big) \\ -(1-c^n) \log\big(\mathcal{F}_{hazard}^n(t|H_F^n)\big), \quad (7)$$

where $\mathcal{F}_{surv}^n$ and $\mathcal{F}_{hazard}^n$ are the survival and hazard functions, as described in Sec. 2.1.

Finally, the total objective function of our framework is a weighted sum of the survival prediction loss $\mathcal{L}_{sur}$ and the mutual contrastive learning loss $\mathcal{L}_{mcl}$ defined in Eq. (6), which is formulated as follows:

$$\mathcal{L}_{total} = \mathcal{L}_{sur} + \lambda \mathcal{L}_{mcl}. \quad (8)$$

where $\lambda$ is the weighting coefficient to balance these two terms.

## 3 Experiments

### 3.1 Experimental Settings

**Datasets.** We train and evaluate our method on three public datasets from The Cancer Genome Atlas (TCGA), i.e., Breast Invasive Carcinoma (BRCA, $N$=956), Lung Adenocarcinoma (LUAD, $N$=453), and Uterine Corpus Endometrial Carcinoma (UCEC, $N$=480). Each dataset includes diagnostic WSIs and ground-truth survival labels.

**Evaluation Metrics.** We adopt the concordance index (CI), a standard metric for survival prediction that quantifies a model's ability to correctly rank predicted risk scores with respect to ground-truth survival labels. In addition, Kaplan-Meier analysis with the Log-rank test is employed to assess the stratification ability between high-risk and low-risk patient groups. To mitigate the impact of data splits, we perform 5-fold cross-validation, and report the mean performance over five folds.

**Implementation Details.** Our network is implemented by Pytorch framework and trained end-to-end on a GeForce GTX 3090 GPUs for 20 epochs with a batch size of 1, using Adam optimizer. The learning rate is set to 2e-4. The ratio $r\%$ in OPL is set to 0.6 and the length of memory queue $B$ in MCL is set to 20. The hyper-parameter $\lambda$ in Eq. (8) is set as 0.01.

## 3.2    Comparisons with State-of-the-Art Approaches

We compare our method against two groups of state-of-the-art (SOTA) approaches: (1) ***Vision-only (V-only) methods***, including ABMIL [6], CLAM [8], TransMIL [7], DSMIL [28], DTFT-MIL [29], and RRT-MIL [5]; and (2) ***Vision-Language (VL) methods***, involving CoOp [30], and VLSA [21]. Specifically, since CoOp is not originally designed for survival prediction, we adapt it by calculating the cosine similarity between visual tokens and language prompts. The top $r\%$ visual tokens, based on cosine similarity values, are selected for predicting risk scores via a prediction layer.

**C-index Comparison.** As shown in Table 1, our HiLa method achieves the state-of-the-art performance with an average CI of 67.1%. Specifically, it surpasses V-only methods by 4.3%-8.2% in overall performance, demonstrating the benefits of integrating language information to enhance traditional MIL pipelines for survival prediction.

**Table 1.** C-index results for different methods on three TCGA datasets. "V" represents the utilization of WSIs and "L" signifies the use of language prompt(s).

| Method | Modality | | Datasets | | | Overall |
|---|---|---|---|---|---|---|
| | V | L | BRCA | LUAD | UCEC | |
| ABMIL [6] | ✓ | | 0.613±0.033 | 0.604±0.043 | 0.618±0.014 | 0.612 |
| CLAM-SB [8] | ✓ | | 0.605±0.062 | 0.586±0.033 | 0.576±0.043 | 0.589 |
| CLAM-MB [8] | ✓ | | 0.611±0.041 | 0.612±0.022 | 0.589±0.023 | 0.604 |
| TransMIL [7] | ✓ | | 0.628±0.015 | 0.626±0.029 | 0.601±0.030 | 0.618 |
| DSMIL [28] | ✓ | | 0.617±0.029 | 0.614±0.035 | 0.647±0.022 | 0.626 |
| DTFT-MIL [29] | ✓ | | 0.612±0.032 | 0.611±0.038 | 0.641±0.032 | 0.621 |
| RRT-MIL [5] | ✓ | | 0.641±0.006 | 0.624±0.029 | 0.618±0.030 | 0.628 |
| CoOp [30] | ✓ | ✓ | 0.608±0.057 | 0.626±0.055 | 0.664±0.023 | 0.633 |
| VLSA [21] | ✓ | ✓ | 0.622±0.021 | 0.622±0.050 | 0.678±0.043 | 0.641 |
| **HiLa (Ours)** | ✓ | ✓ | **0.659±0.044** | **0.643±0.055** | **0.712±0.047** | **0.671** |

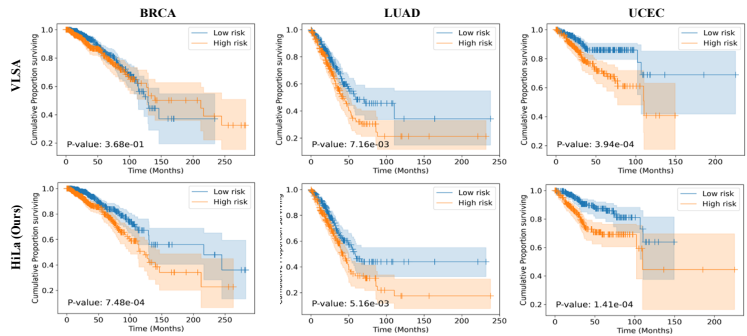

**Fig. 2.** Kaplan-Meier (KM) analysis of the best comparative method and our proposed framework on three datasets. The p-values for Log-rank test are provided in each sub-figure.

**Table 2.** C-index results of the ablation studies on three TCGA datasets.

| ID | Method | | Dataset | | | Overall |
|---|---|---|---|---|---|---|
| | Modules & Descriptions | | BRCA | LUAD | UCEC | |
| (A) | | V-only Baseline | 0.613±0.033 | 0.604±0.043 | 0.618±0.014 | 0.612 |
| (B) | PVLC | (A) + SLP (Cos) | 0.615±0.046 | 0.611±0.035 | 0.637±0.039 | 0.620 |
| (C) | | (A) + MLP (Cos) | 0.616±0.060 | 0.614±0.025 | 0.645±0.034 | 0.625 |
| (D) | | (A) + MLP (OPL) | 0.623±0.017 | 0.617±0.057 | 0.679±0.046 | 0.640 |
| (E) | RVLC | (D) + Region Tokens | 0.644±0.028 | 0.628±0.058 | 0.693±0.042 | 0.655 |
| (F) | CVLC | (E) + CPM | 0.654±0.042 | 0.633±0.047 | 0.702±0.043 | 0.663 |
| (G) | | (F) + MCL (**Ours**) | **0.659±0.044** | **0.643±0.055** | **0.712±0.047** | **0.671** |

Moreover, HiLa also delivers superior performance compared to other VL-based methods. Particularly, HiLa leads the second-best VLSA model by 3.7% on BRCA, 2.1% on LUAD, and 3.4% on UCEC, respectively, confirming its robust capability to exploit survival-related information through vision-language collaboration.

**Kaplan-Meier Analysis.** To further validate the effectiveness of HiLa, we conduct the Kaplan-Meier (KM) analysis. As illustrated in Fig. 2, all patients are stratified into low-risk group (blue) and high-risk group (yellow) based on the median value of predicted risk scores. A Log-rank test is performed to assess the statistical significance of differences between the two risk groups via p-values. Compared to the second-best VLSA, our approach consistently yields p-values below 0.05, suggesting a statistically significant discrimination between the high-risk and low-risk patient groups.

### 3.3 Ablation Studies

To verify the effectiveness of the key components in our HiLa framework, we conduct ablation studies with the following variants. First, ABMIL aggregates patch-level visual tokens, establishing the V-only baseline Model-A. Then, we introduce a single language prompt (SLP) and apply the cosine similarity (Cos) metric to select the top $r\%$ visual tokens, creating Model-B as the baseline for patch-level vision-language collaboration (PVLC). Next, we replace SLP with multiple language prompts (MLP) to form Model-C. Model-D is derived by substituting Cos with the proposed OPL. Building on Model-D, we involve region-level VLC (RVLC) by introducing the selected region-level visual tokens to produce Model-E. Advancing from Model-E, we inject cross-level VLC (CVLC) by progressively incorporating CPM and MCL, yielding Mode-F and Model-G (Ours). The results are provided in Table 2. Particularly, the increasing CI values observed in Models A to D highlight the effectiveness of combining multiple language supervisions and leveraging OPL to exploit discriminative visual tokens associated with diverse survival-associated attributes comprehensively. Moreover, the model achieves superior performance by introducing RVLC (Model-E), indicating the potential of region-level contexts in providing crucial complementary information. In addition, including CPM and MCL (into Models E and F) leads to substantial performance gains, confirming their critical role in fostering hierarchical cooperation.

## 4    Conclusion

In this paper, we have presented a hierarchical vision-language collaboration framework for survival prediction in CPath. To enhance vision-language alignment, we generate multiple survival-related language prompts using LLM. Then, we utilize OPL to encourage comprehensive exploration and identification of discriminative visual features from gigapixel WSIs. Moreover, we also design the CLP and MCL modules, to encourage effective hierarchical cooperation between visual features extracted at patch and region levels. Experimental results have shown our feasibility and superiority.